\documentclass[runningheads]{llncs}
\usepackage{graphicx}

\usepackage{tikz}
\usepackage{comment}
\usepackage{amsmath,amssymb} 
\usepackage{color}

\usepackage[accsupp]{axessibility}  

\usepackage{graphicx}
\usepackage{amsmath}
\usepackage{amssymb}
\usepackage{booktabs}

\usepackage{booktabs}
\usepackage{tabularx}

\usepackage{graphicx}

\usepackage{float}
\usepackage{multirow}
\usepackage{diagbox}

\usepackage{cuted}
\usepackage{capt-of}
\usepackage{pifont}
\usepackage{color}

\usepackage{indentfirst}
\usepackage{hyperref}

\usepackage[capitalize]{cleveref}

\begin{document}
\pagestyle{headings}
\mainmatter
\def\ECCVSubNumber{1289}  

\title{FishFormer: Annulus Slicing-based Transformer for Fisheye Rectification with Efficacy Domain Exploration} 

\author{Shangrong Yang\inst{1} \and
Chunyu Lin\inst{1}\thanks{Corresponding author: cylin@bjtu.edu.cn} \and
Kang Liao\inst{1} \and
Yao Zhao\inst{1}
}
\authorrunning{F. Author et al.}
%
\institute{Institute of Information Science, Beijing Jiaotong University, Beijing Key Laboratory of Advanced Information Science and Network, Beijing, 100044, China 
\email{\{sr\_yang,cylin,kang\_liao,yzhao\}@bjtu.edu.cn}
}

\maketitle

\begin{abstract}
Numerous significant progress on fisheye image rectification has been achieved through CNN. Nevertheless, constrained by a fixed receptive field, the global distribution and the local symmetry of the distortion have not been fully exploited. To leverage these two characteristics, we introduced Fishformer that processes the fisheye image as a sequence to enhance global and local perception. We tuned the Transformer according to the structural properties of fisheye images. First, the uneven distortion distribution in patches generated by the existing square slicing method confuses the network, resulting in difficult training. Therefore, we propose an annulus slicing method to maintain the consistency of the distortion in each patch, thus perceiving the distortion distribution well. Second, we analyze that different distortion parameters have their own efficacy domains. Hence, the perception of the local area is as important as the global, but Transformer has a weakness for local texture perception. Therefore, we propose a novel layer attention mechanism to enhance the local perception and texture transfer. Our network simultaneously implements global perception and focused local perception decided by the different parameters. Extensive experiments demonstrate that our method provides superior performance compared with state-of-the-art methods.
\end{abstract}

\section{Introduction}

Benefiting from the large field of view (FOV), the fisheye camera has become popular in several computer vision tasks such as robot navigation \cite{navigation}, object detection and tracking \cite{tracking}, and motion estimation \cite{motionestimate}. Different from the ideal pinhole model, fisheye cameras distort the incident light to accommodate more content in limited image space, thus they distort the structure as well. The distortion dramatically affects many computer vision algorithms designed on perspective images. Therefore, it is necessary to correct the distortion of the fisheye image.

\begin{figure}[!t]
  \centering
  \includegraphics[scale=0.5]{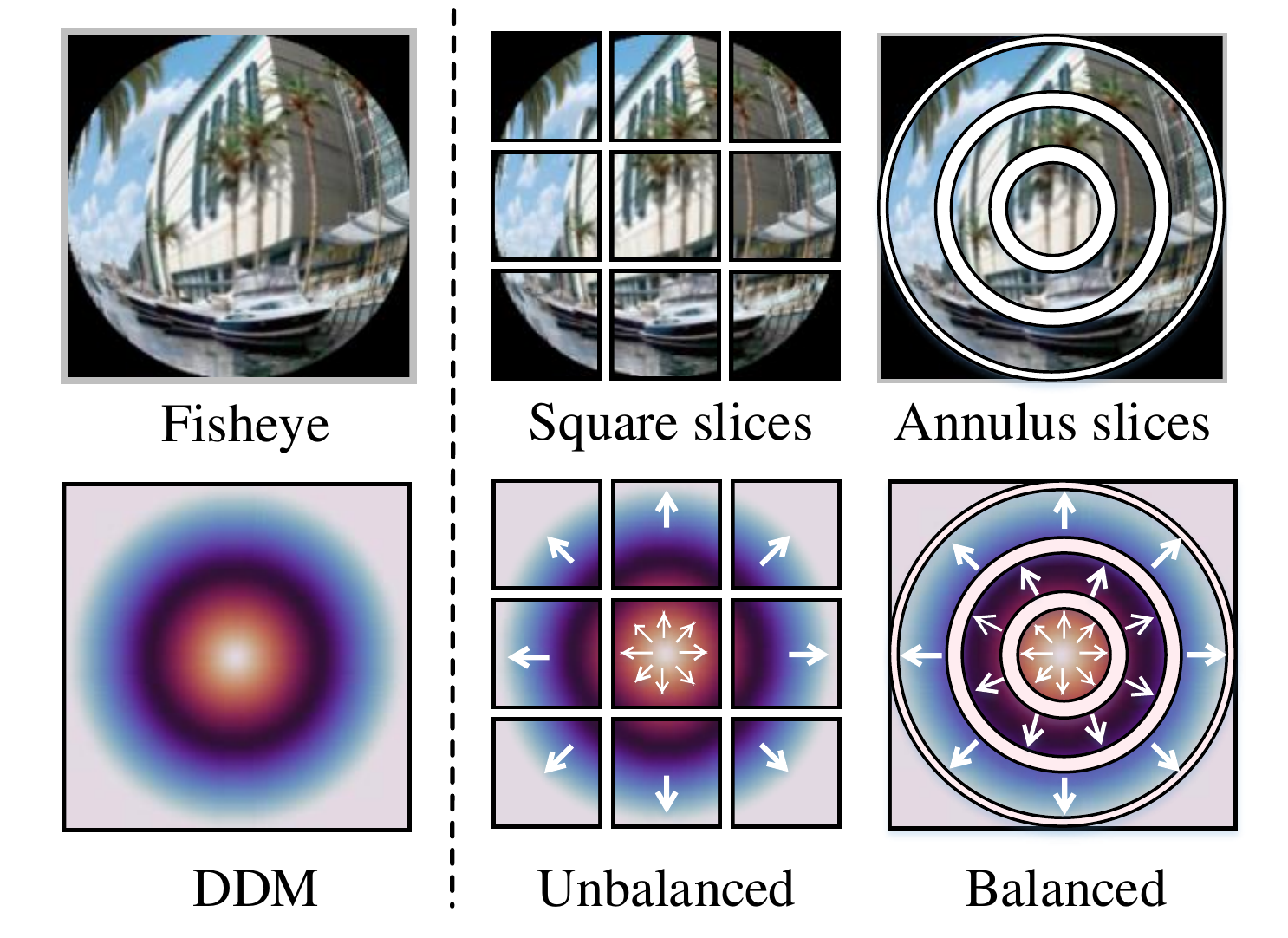}
  \vspace{-0.4cm}
  \caption{
  \label{unb_ddm}
  Given a distortion distribution map (DDM), existing square slicing method (middle) complicates the distortion distribution between patches, thereby increasing the difficulty of network learning. Our learning conducive annulus slicing method (right) ensures the distortion uniformity in a patch and distortion gradualness between different patches.}
  \vspace{-0.6cm}
  \end{figure}

  The distortion in the fisheye image primarily affects the image structure rather than content. Therefore, most previous methods used significant distortion features, such as conics or distorted objects, to help distortion correction. The manual calibration methods \cite{Reimer2013INTCAL1A}\cite{ZHANG1999} can effectively find conics, but it is time-consuming. Self-calibration methods \cite{Dansereau2013DecodingCA}\cite{Stuiver1993Extended1D}\cite{Rui2014Unsupervised} replace artificially searching with automated algorithms, but the found features are not always reliable. To solve these problems, many studies rely on deep neural networks. The regression-based method \cite{Krizhevsky2017ImageNetCW}\cite{Rong2016Radial}\cite{Yin2018FishEyeRecNet}\cite{Xue2019} simplifies the distortion correction to a parameter-prediction problem. Generation-based method \cite{Liao2019}\cite{DDM} directly leverages the generative adversarial network (GAN) to achieve image-to-image correction. Similarly, they use various features, such as edges \cite{DDM}, optical flow \cite{Blind} to enhance the quality of generation. However, these methods have two issues: (1) Convolving the entire image straightforward with unified perception is not conducive to perceiving the global distortion. (2) The radial symmetry of fisheye images is not fully exploited to improve prediction accuracy.

  To handle the problems above, we propose our Fishformer. We improved the Transformer according to the structural characterizations of fisheye images. Note that distortion in fisheye images has high symmetry. But the existing square slicing method would break this symmetry. As shown in Figure \ref{unb_ddm}, it will complicate the range of distortion between patches, thereby increasing the difficulty of global perception. Therefore, we introduce a learning conducive annulus slicing method. The fisheye image is sliced in the annulus to ensure distortion uniformity, facilitating global prediction. By visualizing the efficacy domains of the fisheye parameters, we found different distortion parameters affect relatively fixed regions in an image. \emph{This phenomenon implies that the perception of the local area can further improve the prediction accuracy}. Therefore, we propose a novel layer attention mechanism, which calculates the attention of features between adjacent layers. In this way, the rich texture in the first layer feature will be continuously transferred, thereby enhancing the perception of local areas. Since each patch is in a different area, the confidence of the distortion parameters predicted by each patch should be related to their location. In order to implement focused local perception of different parameters on different regions, we use a set of rough truncated normal distribution probability densities as the weight of patch prediction loss.
  
  To summarize, we make the following contributions:
  \begin{itemize}
    \item We process fisheye images as sequences for the first time and propose a learning conducive annulus slicing method to ensure the uniformity and graduality of regional distortion.
    \item Based on discovered parameters efficacy domains, we propose a novel layer attention mechanism to promote the texture transmission and enhance the focused local perception in different regions.
    \item Our network achieves the perception of global and local texture simultaneously. Extensive experiments on different datasets demonstrate the superiority of our method.
  \end{itemize}
\section{Related Work}
\vspace{-0.1cm}
Distortion will drastically decrease the performance of many perspective image-based algorithms. The distortion rectification can effectively alleviate this problem and thus receive significant attention. The earliest attempt used traditional machine learning methods \cite{Mei2007}\cite{Gasparini2009}\cite{Puig2010}\cite{Zhang2000AFN} to exploit distortion correction. Mei et al. \cite{Mei2007} proposed a method to calculate the radial distortion parameter using a calibration board with coplanar points. It can accurately calculate the parameters but require special equipment and manual participation. Self-calibration methods \cite{Zhang2015}\cite{Barreto2005}\cite{Chander2009SummaryOC}\cite{Geiger2012AutomaticCA} broke down the limitation of human participation. Zhang et al. \cite{Zhang2015} designed a method following the rule that straight lines have to be straight \cite{StraightLine}.  \emph{Detecting the edges and calculating the distortion parameters}. Although the automatic calibration method was more convenient, the detected features were easily affected by the image content, resulting in unstable performance.

With the continued development of deep learning, many researchers considered using deep convolutional networks to correct distortion. Rong et al. \cite{Rong2016Radial} regard distortion correction as a regression task and first introduced Alexnet to perceive distortion and predict parameters. Since their modeling of the fisheye image is straightforward and the parameter is in a relatively small range, the rectification is limited. Yin et al. \cite{Yin2018FishEyeRecNet} proposed a FisheyeRecnet, which is aided by the semantic information to perceive the distortion better. However, the pre-training of the semantic segmentation network further increased the difficulty of training. Similarly, Xue et al. \cite{Xue2019} enhanced their regression network by leveraging the edges detector, which also needs to be pre-trained.

\begin{figure*}[!t]
  \centering
  \includegraphics[scale=0.35]{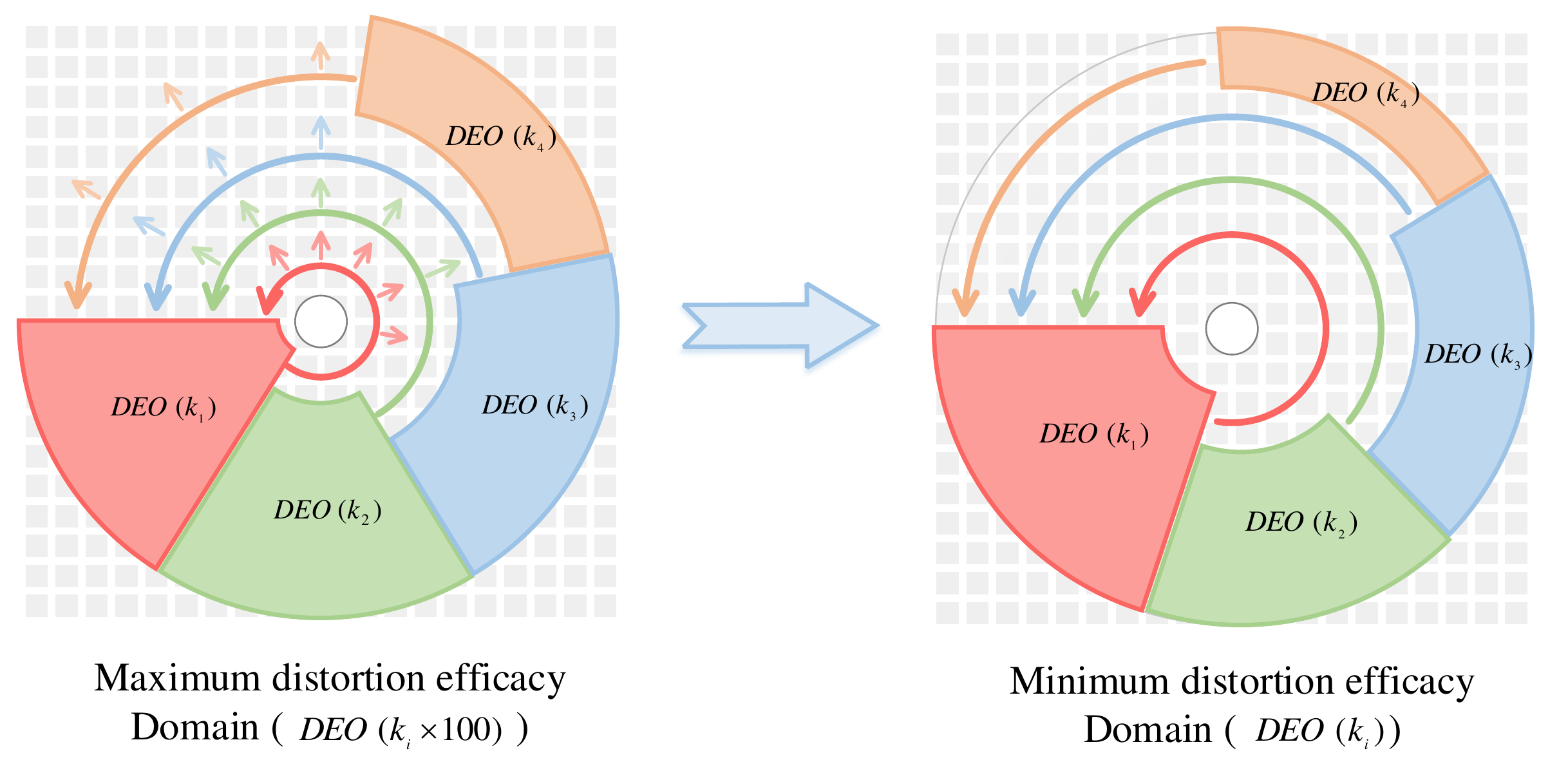}
  \vspace{-0.4cm}
  \caption{
    \label{dis_dom}
  \textbf{Distortion efficacy domain (DEO).} The visualization of the maximum (left) and minimum (right) distortion efficacy domain. The area of the DEO will increase with the value of $k$ and $DEO\left ( k_{1} \right )\geqslant DEO\left ( k_{2} \right )\geqslant DEO\left ( k_{3} \right )\geqslant DEO\left ( k_{4} \right )$.}
  \vspace{-0.5cm}
  \end{figure*}
  
Liao et al. \cite{Liao2019} first considered distortion correction as a generation task. They introduced a simple generative adversarial network (GAN) to learn the distribution difference and achieved one-stage correction. However, naive GAN cannot learn the distribution difference well, so the corrected image has a noticeable artifact. DDM \cite{DDM} improved the correction accuracy by distortion distribution map, but it cannot prevent the content of the corrected image from being changed. Based on this, Yang et al. \cite{PCN} believed that the issue of the ambiguity resides in the transmitted distorted structure, so a network is introduced to isolate the distortion transmission. Although the method of Yang et al. \cite{PCN} has a significant improvement, the accuracy of the appearance flow produced by the self-supervised method is a challenge.
\vspace{-0.4cm}

\section{Fisheye Distortion}
\subsection{Distortion Model}

Since large amount of real fisheye images and their corresponding distortion parameters are difficult to obtain. Using distortion models to synthesize fisheye images has become the mainstream choice. Generally, the commonly used fisheye distortion models is polynomial model \cite{Basu1995}. Suppose the coordinate of an arbitrary point on normal image is $p\left(x,y\right)$. Its corresponding coordinate on fisheye image is ${{p}'\left ({x}',{y}'\right )}$. The Euclidean distance $r_{u}$ from $p\left(x,y\right)$ to the image center $c\left ( x_{0}, y_{0}\right )$ corresponds to the Euclidean distance $r_{d}$ from $p_{d}$ to the distortion center $c_{d}\left ( x_{d}, y_{d}\right )$. They can be expressed as:
\begin{equation}
  r_{u}= \sqrt{\left ( x-x_{0} \right )^{2}+\left ( y-y_{0} \right )^{2}}
\end{equation}
\begin{equation}
  r_{d}= \sqrt{\left ( {x}'-x_{d} \right )^{2}+\left ( {y}'-y_{d} \right )^{2}}
\end{equation}

The polynomial model uses high-order polynomials to fit the complex distortion on fisheye image and can be denoted as:
\begin{equation}
  x=(1+k_{1}r_{d}^{2}+k_{2}r_{d}^{4}+\cdots ){x}'
\end{equation}
\begin{equation}
  y=(1+k_{1}r_{d}^{2}+k_{2}r_{d}^{4}+\cdots ){y}'
\end{equation}

Merge the above formulas when the distortion center $c_{d}\left ( x_{d}, y_{d}\right )$ is the image center. The polynomial model used to describe the relationship between $r_{u}$ and $r_{d}$ can be obtained:
\begin{equation}
  r_{u}=(1+\sum _{i=1}^{n}k_{i}{r_{d}}^{2i})r_{d}
\end{equation}

\subsection{Distortion efficacy domain}
Higher-order polynomials can accurately model real fisheye images. However, the difficulty of network prediction will increase with the number of parameters. Usually, large amount of previous research \cite{Yin2018FishEyeRecNet} \cite{Liao2019} \cite{Xue2019} considered that the four parameters are appropriate choice. Therefore, we also select a four-parameter polynomial model. In addition, for intuitively observing the influence of different parameters on variable areas on the fisheye image, we simplify the polynomial model by treating the image center as the distortion center:
\begin{equation}
  r_{u}=(1+k_{1}r_{d}^{2}+k_{2}r_{d}^{4}+k_{3}r_{d}^{6}+k_{4}r_{d}^{8})r_{d}
\end{equation}

\begin{figure*}[!t]
  \centering
  \includegraphics[scale=.08]{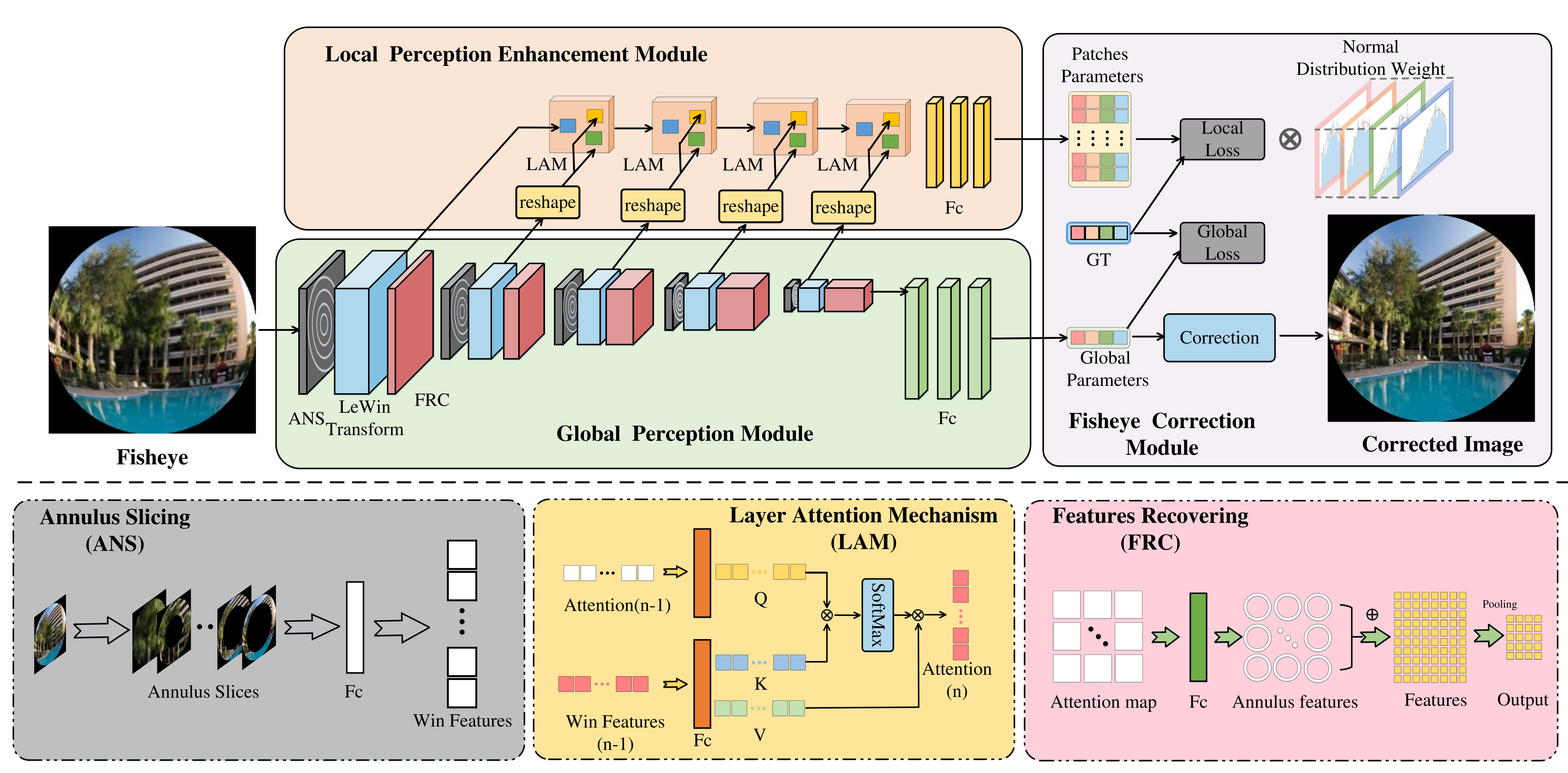}
  \vspace{-0.4cm}
  \caption{
  \label{structure}
  The network architecture of our Fishformer, mainly consists of a global perception module and a local perception enhancement module. In the global perception module, the features are sliced with the annulus slicing method (ANS), then leveraging the LeWin Transformer \cite{wang2021uformer} to calculate the attention. The attention features are recovered (FRC) \emph{to a complete feature}. In the local perception enhancement module, each layer feature is used in layer attention mechanism (LAM) to enhance local perception further.}
  \vspace{-0.4cm}
  \end{figure*}

Notice that the polynomial multiplied by $r_{d}$ contains a linear constant $1$ and a nonlinear term $\sum _{i=1}^{4}k_{i}{r_{d}}^{2i}$. The linear constant represents the scaling transform and does not have impact on distortion. Nonlinear term $\sum _{i=1}^{n}k_{i}{r_{d}}^{2i}$ denotes non-linear transform. To explore the efficacy domain of each nonlinear term, we decompose the formula:
\begin{equation}
\begin{aligned}
  r_{u}=&\left ( \frac{1}{4}+k_{1}{r_{d}}^{2} \right )r_{d}+\left ( \frac{1}{4}+k_{2}{r_{d}}^{4} \right )r_{d}+ \\
  & \left ( \frac{1}{4}+k_{3}{r_{d}}^{6} \right )r_{d}+\left ( \frac{1}{4}+k_{4}{r_{d}}^{8} \right )r_{d}
\end{aligned}
\end{equation}

We set $r(k_{i})=\left ( \frac{1}{4}+k_{i}{r_{d}}^{2i} \right )r_{d}$. Each $r(k_{i})$ is only affected by a linear term and a nonlinear term, and each nonlinear term has only one $k$. To explore the area where each k value works, we only keep $r(k_{i})$ and set the rest to 0. In this way, there are four independent situations:
\vspace{-0.3cm}
\begin{equation}
\begin{aligned}
  &k_{1}\epsilon \left [ 10^{-6}\ \ ,10^{-4} \ \ \right ],r_{u}=\left ( \frac{1}{4} +k_{1}{r_{d}}^{2}\right )r_{d}\\
  &k_{2}\epsilon \left [ 10^{-11},10^{-9}\ \ \right ],r_{u}=\left ( \frac{1}{4} +k_{2}{r_{d}}^{4}\right )r_{d}\\
  &k_{3}\epsilon \left [ 10^{-16},10^{-14} \right ],r_{u}=\left ( \frac{1}{4} +k_{3}{r_{d}}^{6}\right )r_{d}\\
  &k_{4}\epsilon \left [ 10^{-21},10^{-19} \right ],r_{u}=\left ( \frac{1}{4} +k_{4}{r_{d}}^{8}\right )r_{d}
\end{aligned}\label{para_range}
\end{equation}
Among them, please refers to \cite{Liao2019} \cite{Xue2019} for the range of $\left [ k_{1},k_{2},k_{3},k_{4} \right ]$. We use the above four formulas to synthesize fisheye images independently. Then, we provide formula $r_{u}-\frac{1}{4}r_{d}<1$ to divide the distorted region and the distortion-negligible region. The distorted area is called the distortion efficacy domain(DEO). We draw the DEO of each parameter according to the value range of Equation (\ref{para_range}). The DEO of each parameter appears as a 360-degree annular (in Figure \ref{dis_dom} we only color a part of the annular). When we increase the value of $k_{i}$, its corresponding DEO area will increase. The DEO with higher-order coefficients is always smaller than the DEO with lower-order coefficients. The relationship can be expressed as:
\begin{equation}
  DEO\left ( \left ( k_{i} \right )_{max} \right )\geqslant  DEO\left ( k_{i} \right )\geqslant DEO\left ( \left ( k_{i} \right )_{min} \right )
\end{equation}
$DEO\left ( \cdot  \right )$ is the area of the distortion efficacy domain. In the same quantization interval, the higher order of $k$ has the smaller area of the domain. It can be expressed as:
\vspace{-0.2cm}
\begin{equation}
    K_{N}=\sum_{i}^{n}k_{N\_i}
  \end{equation}
\begin{equation}
\begin{aligned}
  DEO\left ( k_{1\_i} \right )\geqslant  &DEO\left ( k_{2\_i} \right )\geqslant \\
  & DEO\left ( k_{3\_i} \right )\geqslant DEO\left ( k_{4\_i} \right )
\end{aligned}
\end{equation}
$K_{N}$ is the range of the N-th $k,N\epsilon \left [ 1,2,3,4 \right ]$, $k_{N\_i}$ is the i-th interval of $K_{N}$. In this way, we can determine the approximate efficacy domains of each $k$, then we design a network that can focus on different parameters on different regions.

\section{Proposed Method}

Most of the existing deep learning methods use convolution to perform local perception but fail to exploit the prior information better. In fact, the distortion structure of the fisheye image has good symmetry. The degree of distortion increase with the radius, showing good progress. Therefore, we process the fisheye image as a sequence. We design our Fishformer based on the existing popular LeWin Transformer \cite{wang2021uformer}, as shown in Figure \ref{structure}. Our network comprises a global perception module and a local enhanced perception module. We first slice the fisheye image in the annulus and send them to the global perception module for feature extraction and parameter prediction. Subsequently, we send the extracted features in each layer to the local enhanced perception module. Calculating the attention between two adjacent layer features maintains local texture transfer and reinforces the local perception.
\vspace{-0.3cm}

\subsection{Global Perception Module}
\subsubsection{Annulus Slicing (ANS).}
The global perception module mainly uses five Fishformer blocks to extract the features and make a global prediction. In the Transformer block, different from the previous methods, we first use equidistant annulus slicing with a width of 8 to replace the traditional square slicing. Square slicing destroys this regularity, leading to inconsistent distortions between patches, thereby increasing the difficulty of network training. In contrast, annulus slicing allows each patch to maintain the same characteristics. The distortion degree in the same patch still remains radial symmetry, and the distortion degree between the different patches maintains an orderly progression. Therefore, the network can learn the distortion characteristics easily. Assuming that the number of slices is $n$, we can transform an image feature of shape $(1, w, h, c)$ into $(n, w, h, c)$ by annulus slicing. Note that annulus slicing causes redundant areas without valid pixels in each patch. Thus, we transform slices into vectors with shape $(n, w*h, c)$ and feed them into a fully connected layer to extract the effective pixels. We get the output with shape $(n, 8*8, c)$ and arrange it into an $8\times8$ patch window with shape $(n, 8, 8, c)$.  
\vspace{-0.4cm}

\subsubsection{Features Recovering (FRC).}
Assisted by the LeWin Transformer structure \cite{wang2021uformer}, each of our Fishformer blocks contains two LeWin Transformer layers to calculate the patch attention. After obtaining the attention with shape $(n, 8, 8, c)$, we need to introduce a fully connected layer for inverse transformation, turning the attention result into the annulus shape with size of $(n, w, h, c)$ and adding them to generate the final attention feature with shape $(1, w, h, c)$. We perform a down-sampling for the attention feature and get the input with shape $(1, w/2, h/2, c)$ for the next block. After five Fishformer blocks, three fully connected layers are used to predict global distortion parameters $P_{gb}$.

We leverage L1 distance to supervise. The distortion parameter ground truth is denoted as $P_{gt}$ and the loss function can be expressed as
\vspace{-0.3cm}
\begin{equation}
  \mathcal{L}_{global}=\sum _{i=1}^{4}\left \| P_{gb}-P_{gt} \right \|_{1}
\end{equation}

\subsection{Local Perception Enhancement Module}
\subsubsection{Layer Attention Mechanism (LAM).}
\begin{figure}[!t]
  \centering
  \includegraphics[scale=0.5]{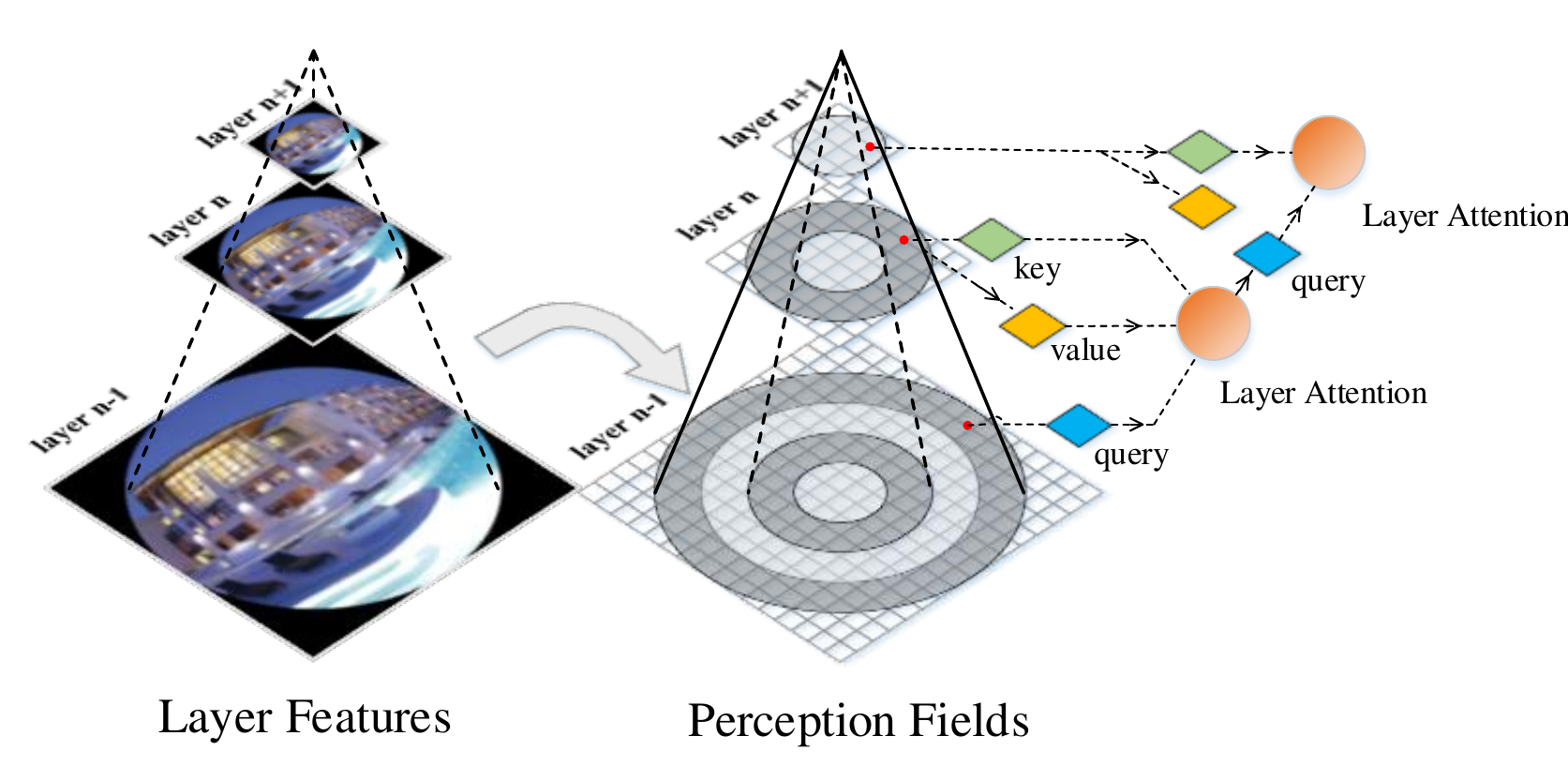}
  \vspace{-0.4cm}
  \caption{
    \label{LAM}
  \textbf{Layer attention mechanism.} By calculating the attention between layer features, the rich texture is continuously passed forward and the relationship between adjacent patches can be quickly constructed.}
  \vspace{-0.5cm}
  \end{figure}

Efficacy domains indicate that different parameters influence different areas. Therefore, the local perception is as important as the global perception. However, Transformer generally focuses on perceiving long-range information and neglecting the local context. Besides, the LeWin Transformer structure down-samples the generated attention features, which will cause the loss of local texture. To solve these problems, we introduce a novel layer attention mechanism. We use the window patches with a size of $(n, 8, 8, c)$ in the first transformer block as query, and the window patches in the next layer with shape $(n/2, 8, 8, c*2)$ as key and value to calculate the attention map. Noticing that the shapes of query and key are different, we resize key and value to the shape of query as the shape $(n, 8, 8, c)$ to calculate the attention. After obtaining the attention map, we resize it to $(n/2, 8, 8, c*2)$. Leveraging it as a new query and the subsequent window patches with shape $(n/4, 8, 8, c*4)$ as key and value, we can calculate the new attention map. In this way, the intense texture detail in the first Transformer block can be continuously forward passed. Compared with LeWin Transformer, the layer attention mechanism can fully exploit the relationship between layer features. The relationship between adjacent patches can be quickly constructed and achieve progressive perception, as shown in Figure \ref{LAM}. Our purpose is to enhance the local perception. Therefore, for each patch, the network needs to predict corresponding distortion parameters. We feed the final attention map into three fully connected layers and obtain multiple patches parameters.

\vspace{-0.5cm}
\subsubsection{Loss Function of Patch Parameters based on Normal Distribution Confidence.}
In the local perception enhancement module, the network locally perceives each patch and then predicts their corresponding distortion parameters. However, each patch of the fisheye image is variably affected by distortion parameters. It means that perceiving different patches assists the network to produce an accurate prediction for some of the parameters. Considering these characteristics, we introduce different truncated normal distribution densities for different parameters as their confidence. There are four truncated normal distribution density functions for our four distortion parameters. We artificially set their variance $\sigma$ to be the same, denoted as $\sigma=1$. Their mean $\mu$ is set to $\left [ -1, -0.5, 0.5, 1  \right ]$ and truncated interval is set to $\left [ -2, 2 \right ]$. The normal distribution density function and parameters confidence is as follows:

\begin{equation}
  f\left ( x,\sigma ,\mu  \right )=\frac{1}{\sqrt{2\pi }\sigma }e^{-\frac{1}{2}\left ( \frac{x-\mu }{\sigma } \right )^{2}}
\end{equation}
\begin{equation}
  C\left ( k_{i} \right )=f\left ( x,1 ,\mu_{i}  \right ),\mu _{i}\epsilon \left [ -1, -0.5, 0.5, 1 \right ]
\end{equation}

After obtaining the weight of the patch predicted parameters $C_{t}\left ( k_{i} \right )$, we calculate a weighting L2 distance for patch predicted parameters $P_{pt}$ and ground truth $P_{gt}$ as the loss of local enhancement perception module. The process can be expressed as:
\begin{equation}
  \mathcal{L}_{local}=\sum_{i=1}^{4}\sqrt{\frac{\sum _{t=1}^{n}C_{t}\left ( k_{i} \right )\left [ P_{pt}\left ( k_{i} \right ) - P_{gt}\left ( k_{i} \right ) \right ]^{2}}{\sum _{t=1}^{n}C_{t}\left ( k_{i} \right )}}
\end{equation}
Where n is the number of patches.

\vspace{-0.2cm}
\subsection{Fisheye Correction Module}
After obtaining the predicted distortion parameters, the network has been able to achieve regular training. However, it is not a complete end-to-end process from the fisheye image to the corrected image. Therefore, we use the predicted distortion parameters to pre-correct the fisheye image. 

The pixels of the fisheye image and the standard image are not a one-to-one correspondence. There are many holes in the pre-corrected image. We use a moving average filtering method to fill in missing pixels. Specifically, we first calculate a mask $M$ from the pre-corrected image $I_{p}$. We leverage moving average filtering on the pre-corrected image to generate a blurred corrected image $I_{b}$. Then we take out the filled pixels through a mask and add them to the pre-corrected image to get the final corrected image $I_{c}$. The inpainting process can be expressed by the following formula:
\vspace{-0.1cm}
\begin{equation}
  I_{c}=M\cdot I_{p}+\left ( 1-M \right )\cdot I_{b}
\end{equation}

After obtaining the corrected image, we calculate the image Loss to help the network training. There are two advantages: (1)The gap between the image domains is relatively small, which benefits the network to estimate parameters quickly. (2)A complete end-to-end training from fisheye image to corrected image is realized.

We use L2 distance as the image loss to supervise the quality of the corrected image. The calculation can be expressed as:
\vspace{-0.2cm}
\begin{equation}
  \mathcal{L}_{image}=\left \| I_{c}-I_{gt} \right \|_{2}
\end{equation}

Finally, we construct our training loss function by summing together all losses as follows:
\begin{equation}    
  \mathcal{L}=\mathcal{L}_{global}+\mathcal{L}_{local}+\mathcal{L}_{image}
\end{equation}
\section{Experiments}
\subsection{Dataset Generation and Experiment Details}
We select two completely different types of datasets to verify the effectiveness of our method. (1)Place2 dataset \cite{Places2}, which contains more than 10 million images, covering 400 indoor and wild scenes. (2)CelebA dataset \cite{celebA}, which includes 2 million face images. We randomly select 44k images from these two datasets as our original images. Since our method requires annulus slicing according to the distortion distribution, and most real fisheye images are circular. Therefore, we crop the image to be circular instead of square in previous methods \cite{Blind}\cite{Rong2016Radial}\cite{Liao2019}\cite{DeepCalib}\cite{DDM}\cite{PCN}. We leverage the four-parameters polynomial model to transform the normal image, generating fisheye images with $128\times128$ resolution. The value of each parameter is randomly selected in the range described in Equation (\ref{para_range}) for data augmentation. 40K synthetic fisheye images are randomly selected for training, leaving 4k images for testing. For the Place2 and CelebA datasets, we train two different models separately. Considering a robust rectification model, it should not attach \emph{extreme} to the images content but focus on the structure. We thereby exchange the training and test on the Place2 and the CelebA. For example, we evaluate the training model from Places2 on the CelebA test set to verify our network performance and vice versa.

\begin{table}[!t]
  \tiny
  \centering
  \renewcommand\arraystretch{1.1}
  \caption{Comparison between the proposed method and the state-of-the-art.}
  \vspace{-0.2cm}
  \label{result table}
    \begin{tabular}{lccccccccccc}
      \hline
    \multicolumn{1}{c}{\multirow{2}[4]{0.5cm}{Method}} & \multicolumn{5}{c}{\textbf{Places2}}           &       & \multicolumn{5}{c}{\textbf{CelebA}} \\
\cmidrule{2-6}\cmidrule{8-12}          & \multicolumn{1}{p{0.9cm}<{\centering}}{PSNR $\uparrow$} & \multicolumn{1}{p{0.8cm}<{\centering}}{SSIM $\uparrow$} & \multicolumn{1}{p{1.2cm}<{\centering}}{MS-SSIM $\uparrow$} & \multicolumn{1}{p{0.7cm}<{\centering}}{FID $\downarrow$} & \multicolumn{1}{p{1.3cm}<{\centering}}{CW-SSIM $\uparrow$} &       & \multicolumn{1}{p{0.9cm}<{\centering}}{PSNR $\uparrow$} & \multicolumn{1}{c}{SSIM $\uparrow$} & \multicolumn{1}{c}{MS-SSIM $\uparrow$} & \multicolumn{1}{p{0.7cm}<{\centering}}{FID $\downarrow$} & \multicolumn{1}{p{1.3cm}<{\centering}}{CW-SSIM $\uparrow$} \\
\hline
    Blind \cite{Blind}&  14.71      & 0.4716      &  0.5487     &   189.5     &  0.6779     &       &   17.19     &  0.6101     &  0.6645     &   119.4     & 0.7759 \\
    DCCNN \cite{Rong2016Radial}&   15.17    &  0.4760     &  0.3668     &  190.8     &   0.6874    &       &   15.46    &  0.5697     &  0.4311     &  66.4     & 0.7193 \\
    DRGAN \cite{Liao2019}&   17.75    &  0.5558     &   0.7174    &  164.9     &   0.7586    &       &  18.22     &  0.6551     &  0.7563     &   129.9    & 0.8022 \\
    RPC \cite{RPC}  &   19.74    &  0.5497     &   0.6323     &  167.6     &   0.7687    &       &   19.47    &  0.6854     &  0.7332     &   75.7     & 0.8228 \\
    DeepCalib \cite{DeepCalib}&  20.84     &  0.6868     &  0.7722     &   69.7     &  0.8537     &       &  18.03     &  0.7138     &  0.7449     &  140.1     &  0.8177 \\
    DDM \cite{DDM}  &  24.69     &  0.7952     &    0.9173    &  79.5     & 0.9205      &       & 26.53      &   0.8746     &   0.9317    &   52.0    &   0.9374\\
    PCN \cite{PCN}  &    25.10    &   0.8152     &   0.9178    &   \textbf{65.8}     &  0.9323     &       &  27.34     &   0.8844    &  0.9466     &    39.2    &   0.9534\\
    Ours(FishFormer) &  \textbf{25.43}     &  \textbf{0.8412}     &  \textbf{0.9411}     & 73.4      &   \textbf{0.9388}    &       &  \textbf{27.62}     &  \textbf{0.9071}     &  \textbf{0.9595}     &  \textbf{22.3}     & \textbf{0.9566} \\
    \hline
    \end{tabular}%
  \label{tab:addlabel}%
  \vspace{-0.2cm}
\end{table}%

\begin{table}[!t]
  \tiny
  \centering
  \renewcommand\arraystretch{1.1}
  \caption{Cross-testing between our method and a part of the-state-of-art methods.}
  \vspace{-0.2cm}
  \label{exchange_result}
    \begin{tabular}{lccccccccccc}
      \hline
    \multicolumn{1}{c}{\multirow{2}[4]{0.5cm}{Method}} & \multicolumn{5}{c}{\textbf{Places2 $\rightarrow$ CelebA}}           &       & \multicolumn{5}{c}{\textbf{CelebA $\rightarrow$ Places2}} \\
\cmidrule{2-6}\cmidrule{8-12}          & \multicolumn{1}{p{0.9cm}<{\centering}}{PSNR $\uparrow$} & \multicolumn{1}{p{0.8cm}<{\centering}}{SSIM $\uparrow$} & \multicolumn{1}{p{1.2cm}<{\centering}}{MS-SSIM $\uparrow$} & \multicolumn{1}{p{0.7cm}<{\centering}}{FID $\downarrow$} & \multicolumn{1}{p{1.3cm}<{\centering}}{CW-SSIM $\uparrow$} &       & \multicolumn{1}{p{0.9cm}<{\centering}}{PSNR $\uparrow$} & \multicolumn{1}{c}{SSIM $\uparrow$} & \multicolumn{1}{c}{MS-SSIM $\uparrow$} & \multicolumn{1}{p{0.7cm}<{\centering}}{FID $\downarrow$} & \multicolumn{1}{p{1.3cm}<{\centering}}{CW-SSIM $\uparrow$} \\
\hline

    Blind \cite{Blind}&  14.89      &  0.5742      &  0.6266     &    171.5     &  0.7439     &       &   15.93     &  0.4800     &   0.5686     &   222.9     & 0.7048 \\
    DCCNN \cite{Rong2016Radial}&   14.91    &  0.5553    &  0.3978     & 76.9     &   0.7051    &       &   15.31    &  0.4792    & 0.3757     &  186.2     & 0.6913 \\
    DRGAN \cite{Liao2019}&   17.61    &  0.5921    &   0.7281    &  127.1     &   0.7771    &       &  15.13    &  0.4672     &  0.5501    &   298.6    & 0.6765 \\
    PCN \cite{PCN}  &    27.68    &   0.8963     &   0.9523    &   25.3    &  0.9595     &       &  22.57     &   0.7235   &  0.8377     &    111.4    &   0.8866\\
    Ours(Fishformer) &  \textbf{27.89}     &  \textbf{0.9185}     &  \textbf{0.9685}     & \textbf{22.4}     &   \textbf{0.9614}    &       &  \textbf{23.00}     &  \textbf{0.7509}     &  \textbf{0.8546}     &  \textbf{81.6}     & \textbf{0.8953} \\
    \hline
    \end{tabular}%
  \label{tab:addlabel}%
  \vspace{-0.5cm}
\end{table}%

\vspace{-0.3cm}
\subsection{Quantitative Comparison}
To measure our method performance, we compared with state-of-the-art methods including Blind \cite{Blind}, DCCNN \cite{Rong2016Radial}, DRGAN \cite{Liao2019}, RPC \cite{RPC}, DeepCalib \cite{DeepCalib}, DDM \cite{DDM}, PCN \cite{PCN}. Deep-learning methods \cite{Blind}\cite{Rong2016Radial}\cite{Liao2019}\cite{RPC}\cite{DDM}\cite{PCN} are retrained on our circular dataset except \cite{DeepCalib}. For DeepCalib \cite{DeepCalib} with a special model, panoramic images are used to generate fisheye images and corresponding ground truth. However, our dataset has no corresponding panoramic images. Therefore, we did not retrain the DeepCalib \cite{DeepCalib}. We use its regression pretrained model to test our circular images. To improve fairness, we scale the corrected images with different radius as candidate results. Calculating the performance for candidate results, the maximum value is taken as the final performance.

We leverage PSNR (Peak Signal to Noise Ratio), SSIM (Structural Similarity), MS-SSIM (Multiscale structural similarity) \cite{MS-SSIM}, FID (Frechet Inception Distance) \cite{FID}, and CW-SSIM (Complex wavelet structural similarity) \cite{CW_SSIM} to evaluate the performance. 

\begin{figure*}[!t]
  \centering
  \includegraphics[scale=.112]{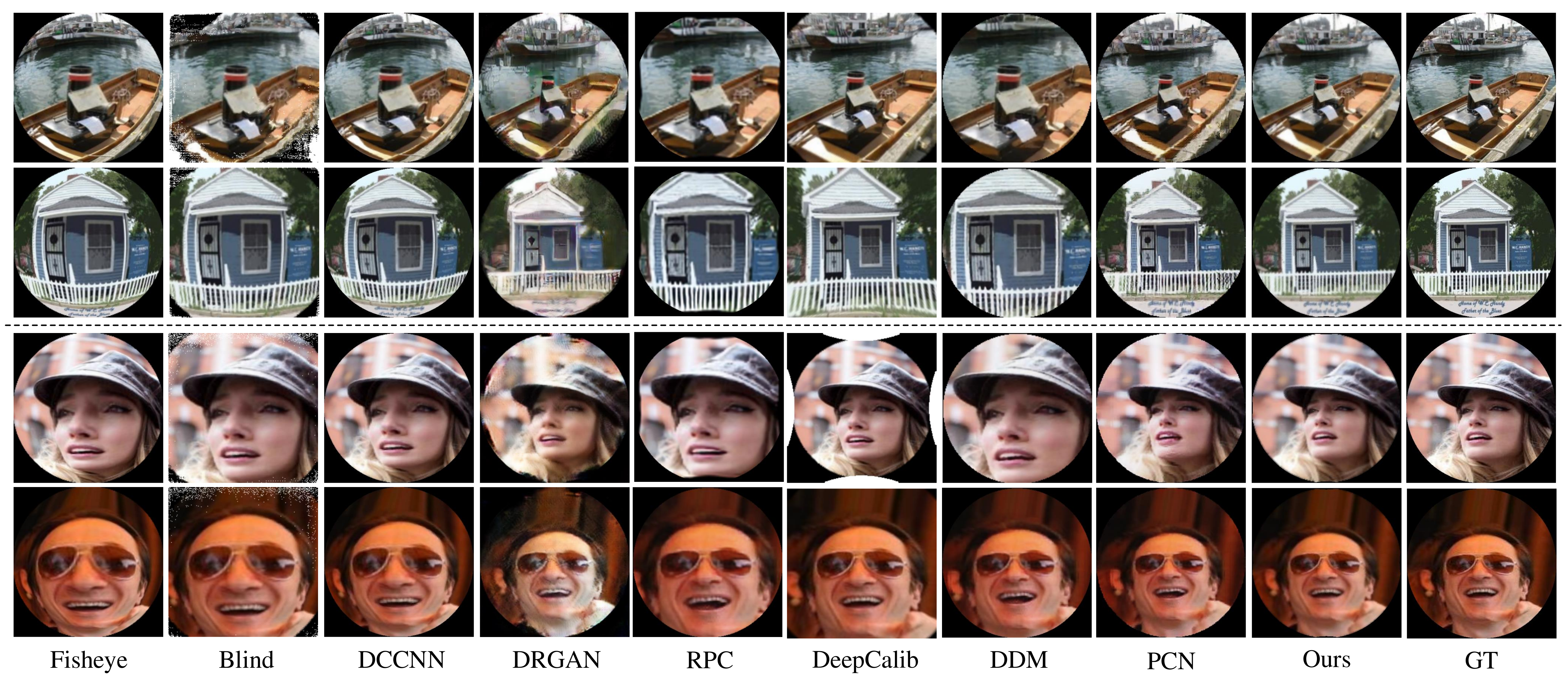}     
  \vspace{-0.8cm}
  \caption{
  \label{result1}
  \textbf{Qualitative Comparison results.} From left to right, we demonstrate the rectification results of Blind \cite{Blind}, DCCNN \cite{Rong2016Radial}, DRGAN \cite{Liao2019}, RPC \cite{RPC}, DeepCalib \cite{DeepCalib}, DDM \cite{DDM}, PCN \cite{PCN}, and our method. Our comparative experiments are tested on synthetic Places2 dataset \cite{Places2} (top two rows) and CelebA \cite{celebA} dataset (bottom two rows), respectively.}
  \vspace{-0.55cm}
  \end{figure*}

  \begin{figure*}[!t]
    \centering
    \includegraphics[scale=.12]{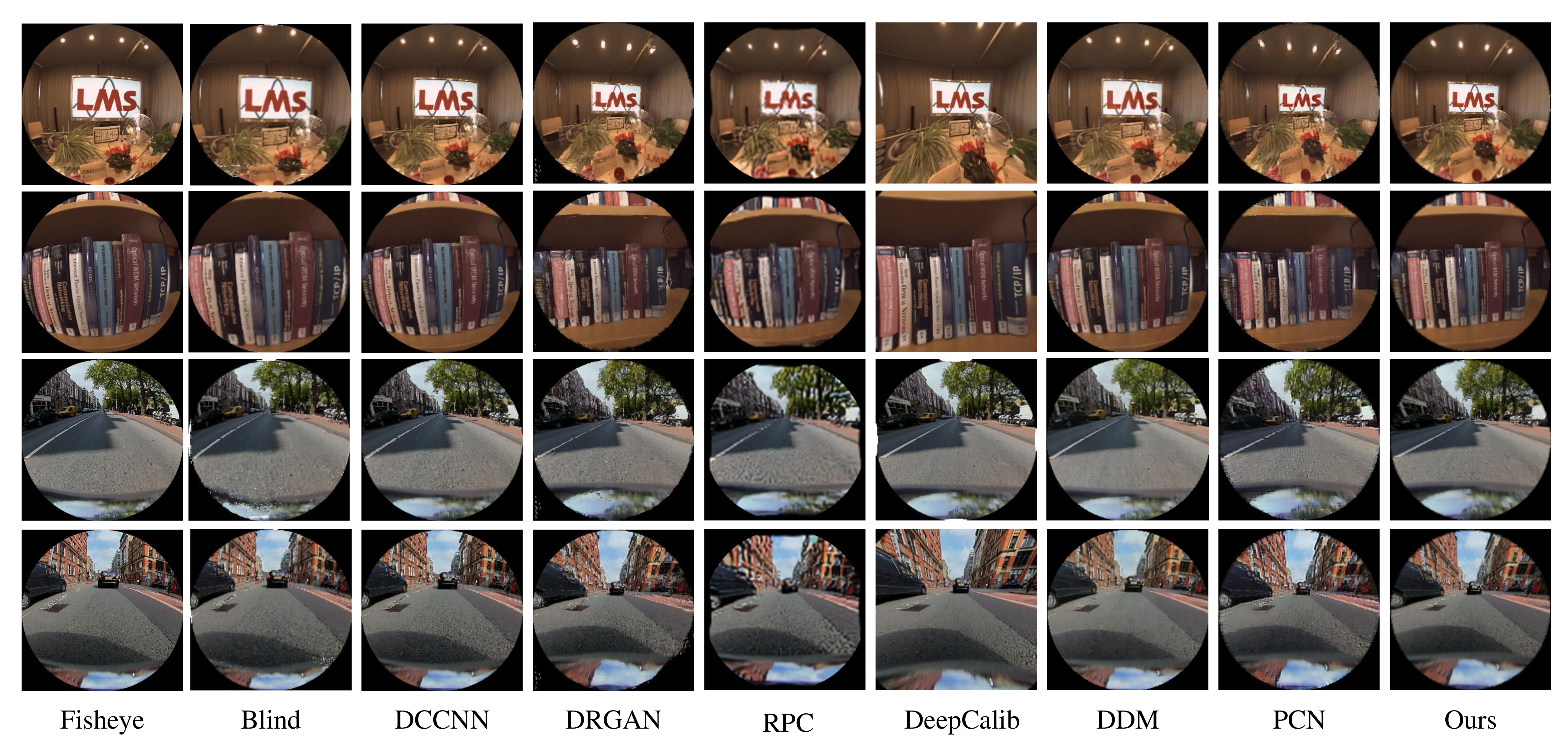}  %
    \vspace{-0.8cm}
    \caption{
    \label{real_result}
    \textbf{Rectification for real fisheye images.} We test the state-of-the-art methods on different real fisheye datasets, including LMS fisheye dataset \cite{LMS} (top) and Woodscape fisheye dataset \cite{woodscape} (bottom).}
    \vspace{-0.6cm}
    \end{figure*}

We test on the Place2 dataset and CelebA dataset respectively, and also conduct cross-testing. The experiment results are shown in Table \ref{result table}. In general, the test results on the CelebA \cite{celebA} are better than those on the Places2 \cite{Places2}. The complexity of the images on CelebA \cite{celebA} is smaller than that on Places2 \cite{Places2}, so it is easier to correct. Limited by the simple model, the performance of \cite{Blind}\cite{Rong2016Radial}\cite{Liao2019} is worse than \cite{RPC}. \cite{DeepCalib} is better than \cite{RPC}, because its model is closer to the real fisheye images. \cite{DDM} and \cite{PCN} utilize Generative Adversarial Networks (GAN) to assist correction with more features, so they lead to a more significant performance boost. As can be seen, the PCN \cite{PCN} is the best deep learning method, but the performance is still lower than our method. When exchanging training models for testing, the model trained on Places2 \cite{Places2} can achieve better performance on CelebA \cite{celebA}. In contrast, the performance will drop when testing the model trained on CelebA. This phenomenon is because the data complexity of Places2 \cite{Places2} is greater than that of CelebA \cite{celebA}, so the trained model can be easily applied to low-complexity data. Nevertheless, our approach is still better than PCN \cite{PCN}, as shown in Table \ref{exchange_result}.

\begin{figure*}[!t]
  \centering
  \includegraphics[scale=.15]{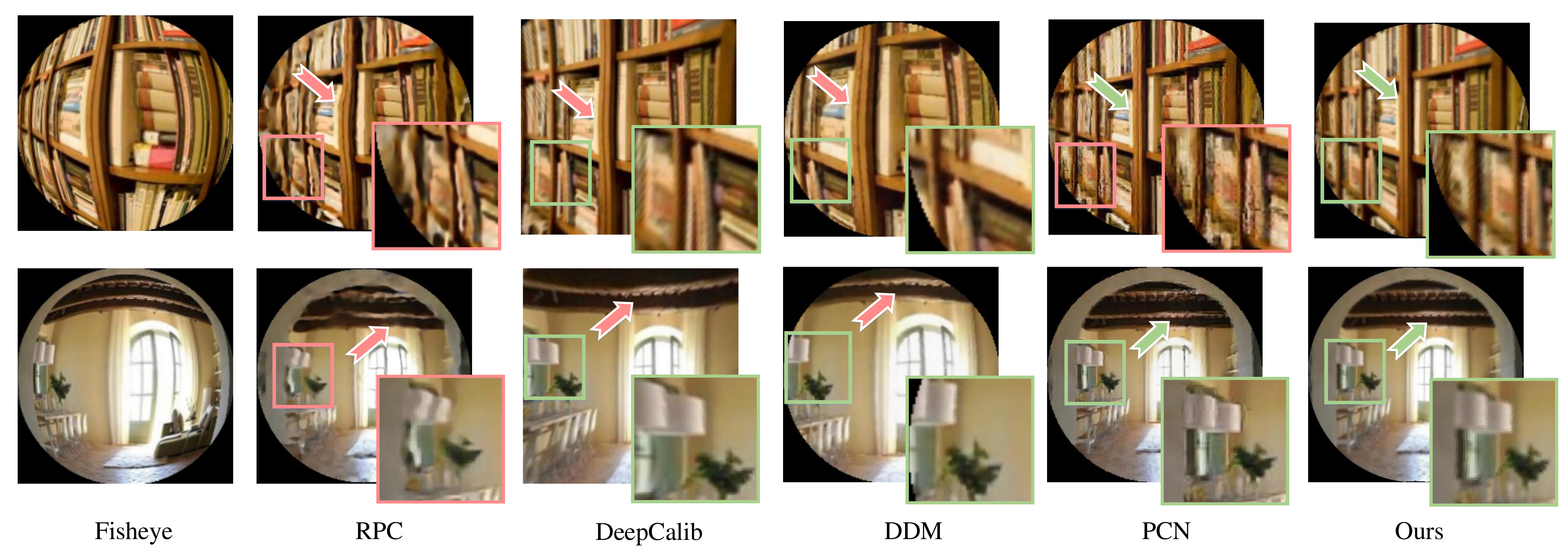}
  \vspace{-0.3cm}
  \caption{
  \label{special_result}
  \textbf{Further comparison in rectification quality.} The arrows and rectangular boxes highlight the corrected structure and content, respectively. Colors represent different qualities: reasonable quality (green), and poor quality (pink).}
  \vspace{-0.6cm}
  \end{figure*}

\subsection{Qualitative Comparison}
We visualized the calibration results of all comparison methods and our method. The synthetic images corrected results are shown in Figure \ref{result1}. The real fisheye images rectified results including LMS dataset \cite{LMS} and Woodscape dataset \cite{woodscape} are shown in Figure \ref{real_result}. 

The correction effect of \cite{Blind}\cite{Rong2016Radial} are not significant because their model is relatively simple. \cite{Liao2019} can correct the structure well, but it will produce obvious artifacts. The two-stage correction of \cite{RPC} further reduce the artifacts. However, the structure correction is still inaccurate. The structural correction of \cite{DeepCalib} is accurate, while the effect varies according to the different images. \cite{DDM} and \cite{PCN} are very accurate in structure correction. Nevertheless, there are some local artifacts caused by the characteristics of the GAN. In general, the generation-based method \cite{Blind}\cite{Liao2019}\cite{DDM}\cite{PCN}\cite{RPC} is better in structure correction, but the generated images have an artifact to some extent. Compared with the generation-based method, the regression method \cite{Rong2016Radial}\cite{DeepCalib} only needs to warp and interpolate the original image according to the predicted parameters. There is less artifact in the central area for most preserved original pixels. Methods \cite{Rong2016Radial}\cite{DeepCalib} do not have the problem of artifact content, but the structure correction is deficient. In contrast, our corrected results achieve quite accurate structure, as shown in Figure \ref{special_result}.

\vspace{-0.3cm}
\subsection{Ablation study}
To further analyze our network, we introduce many additional experiments to exploit the improvements from different key components, including the slicing method, combination of layer attention mechanism, and the network depth.

\vspace{-0.4cm}
\subsubsection{Slicing method and number of patches}
To prove the effectiveness of our proposed slicing method, we use traditional square slicing and annulus slicing to conduct experiments. Besides, we also verify the impact of the patches number. Except for the slicing method and the number of patches, we keep the remaining settings consistent. The experiment results are shown in Table \ref{different slicing}.

The performance without using the slicing method (patch is 1) is worse than using the slicing method. Compared with square slicing, the performance of annulus slicing is always better regardless of the number of patches. This performance effectively proves that annulus slicing can benefit the network to learn the distortion distribution. By using annulus slicing, as the patches number increases, the network achieves optimal performance when the number is 16. It demonstrates that the patches number is not the more the better. Training difficulty will increase with the patches number.

\vspace{-0.4cm}
\subsubsection{Combination of layer attention mechanism}
In this paper, we propose a novel layer attention mechanism. To verify its effectiveness, we removed it and only used the global perception module to predict the distortion parameters. The results are shown in Table \ref{using LAM}. It can be seen that our layer attention mechanism can bring certain improvements.

\begin{table*}[!t]
  \scriptsize
  \centering
  \caption{Result of using different slicing methods and different number of patches.}
  \vspace{-0.3cm}
  \label{different slicing}
    \begin{tabular}{p{1.5cm}<{\centering}p{2.0cm}<{\centering}p{2.0cm}<{\centering}p{2.0cm}<{\centering}p{2.0cm}<{\centering}p{2.0cm}<{\centering}}
    \toprule
    Square/Annulus   & \multicolumn{1}{c}{PSNR $\uparrow$} & \multicolumn{1}{c}{SSIM $\uparrow$} & \multicolumn{1}{c}{MS-SSIM $\uparrow$} & \multicolumn{1}{c}{FID $\downarrow$} & \multicolumn{1}{c}{CW-SSIM $\uparrow$} \\
    \midrule
    1     & 22.96/22.96    &  0.7440/0.7440     &  0.8539/0.8539    &  81.7/81.7    & 0.8933/0.8933 \\
    4     &  23.41/23.81     &  0.7568/0.7786     &   0.8711/0.8891    &  82.3/82.1    
     &  0.9031/0.9112 \\
    16    &  24.11/\textbf{25.34}     &  0.7941/\textbf{0.8314}     & 0.9023/\textbf{0.9326}    &  79.8/\textbf{74.1}     & 0.9176/\textbf{0.9353} \\
    64    &  24.22/24.75     &  0.7914/0.8149     &  0.8975/0.9224     &  79.6/78.9     & 0.9166/0.9271 \\
    \bottomrule
    \end{tabular}%
  \label{tab:addlabel}%
  \vspace{-0.2cm}
\end{table*}%

\begin{table}[!t]
  \footnotesize
  \setlength\tabcolsep{3pt}
  \centering
  \caption{Performance of using layer attention mechanism.}
  \vspace{-0.3cm}
  \label{using LAM}
    \begin{tabular}{cccccc}
    \toprule
          & \multicolumn{1}{c}{PSNR $\uparrow$} & \multicolumn{1}{c}{SSIM $\uparrow$} & \multicolumn{1}{c}{MS-SSIM $\uparrow$} & \multicolumn{1}{c}{FID $\downarrow$} & \multicolumn{1}{c}{CW-SSIM $\uparrow$} \\
    \midrule
    W LTM &  \textbf{25.43}     &  \textbf{0.8412}     &  \textbf{0.9411}     &  \textbf{73.4}     & \textbf{0.9388} \\
    W/O LTM &  25.34     &  0.8314     &  0.9326     &  74.1     & 0.9353 \\
    \bottomrule
    \end{tabular}%
  \label{tab:addlabel}%
  \vspace{-0.2cm}
\end{table}%

Subsequently, we explored the combination of layer attention mechanism. By exploring the combination of q, k, v, we search for optimal performance. The search results are shown in Table \ref{combinations}. We found that when the first layer features are used as q, the obtained performance is the best. This combination can retain more local texture so that the network pays more attention to local perception.

\begin{table}[!t]
  \footnotesize
  \centering
  \caption{Performance of different combinations.}
  \vspace{-0.3cm}
  \setlength\tabcolsep{3pt}
  \label{combinations}
    \begin{tabular}{ccccccc}
    \toprule
    \multicolumn{1}{c}{Layer I} & \multicolumn{1}{c}{Layers II-V} & \multicolumn{1}{c}{PSNR $\uparrow$} & \multicolumn{1}{c}{SSIM $\uparrow$} & \multicolumn{1}{c}{MS-SSIM $\uparrow$} & \multicolumn{1}{c}{FID $\downarrow$} & \multicolumn{1}{c}{CW-SSIM $\uparrow$} \\
    \midrule
    q     & kv    &  25.43     & \textbf{0.8412}      &  \textbf{0.9411}     &  \textbf{73.4}    & \textbf{0.9388} \\
    k     & qv    &  \textbf{25.53}     & 0.8396      &  0.9386     &  74.0    & 0.9384 \\
    v     & qk    &  25.26     &  0.8279     &  0.9309     &  75.1     & 0.9338 \\
    kv    & q     &  25.23     & 0.8334      &  0.9358     &  75.2      & 0.9347 \\
    qv    & k     &  25.08     & 0.8281      &  0.9302     &  74.8     & 0.9329 \\
    qk    & v     &  25.09     &  0.8282     &  0.9302     &  74.9     & 0.9329 \\
    \bottomrule
    \end{tabular}%
  \label{tab:addlabel}%
  \vspace{-0.2cm}
\end{table}%

In addition, to explore whether the local perception is necessary participate in global perception, we deliver the feature in the local perception enhancement module to the global perception module on each layer ($w \ deliver$) and get the experiment results in Table \ref{way_LAM}. The improvement is not apparent. We also conduct additional exploration on layer attention mechanism. Using skip layer features ($w \ skip$) and layer by layer features ($w/o \ skip$) to calculate attention separately. The results are shown in Table \ref{way_LAM}. The layer-by-layer ($w/o \ skip$) result is better than the skip layer ($w \ skip$) result. It demonstrates that layer-by-layer attention can perceive local texture more meticulously.

\begin{table}[!t]
  \footnotesize
  \setlength\tabcolsep{3pt}
  \centering
  \caption{Performance of way of layer attention mechanism.}
  \vspace{-0.3cm}
  \label{way_LAM}
    \begin{tabular}{cccccc}
    \toprule
          & \multicolumn{1}{c}{PSNR $\uparrow$} & \multicolumn{1}{c}{SSIM $\uparrow$} & \multicolumn{1}{c}{MS-SSIM $\uparrow$} & \multicolumn{1}{c}{FID $\downarrow$} & \multicolumn{1}{c}{CW-SSIM $\uparrow$} \\
    \midrule
    $w \ deliver$ &  25.43     &   0.8412    &  0.9404     & \textbf{72.2}      & 0.9376 \\
    $w/o \ deliver$ &  \textbf{25.43}      &  \textbf{0.8412}     &  \textbf{0.9411}     &  73.4     & \textbf{0.9388} \\
    \midrule
    $w \ skip$ &  25.43     &  \textbf{0.8412}     &  \textbf{0.9411}      &  \textbf{73.4}     & \textbf{0.9388}  \\
    $w/o \ skip$ & \textbf{25.45}      &  0.8378     &  0.9384     &  73.7     & 0.9376 \\
    \bottomrule
    \end{tabular}%
  \label{tab:addlabel}%
  \vspace{-0.4cm}
\end{table}%

 \begin{table}[!t]
  \footnotesize
  \centering
  \caption{Performance of different depth.}
  \vspace{-0.3cm}
  \label{different depth}
    \begin{tabular}{cccccc}
    \toprule
    num   & \multicolumn{1}{c}{PSNR $\uparrow$} & \multicolumn{1}{c}{SSIM $\uparrow$} & \multicolumn{1}{c}{MS-SSIM $\uparrow$} & \multicolumn{1}{c}{FID $\downarrow$} & \multicolumn{1}{c}{CW-SSIM $\uparrow$} \\
    \midrule
    3     &  22.98    & 0.7485      & 0.8545     &  82.2     & 0.8943  \\
    4     &  25.06     & 0.8208      & 0.9256      &  75.4     & 0.9306 \\
    5    &   \textbf{25.34}    &  \textbf{0.8314}     &  \textbf{0.9326}     &  \textbf{74.1}     & \textbf{0.9353} \\
    6    &   25.14    & 0.8294      &  0.9322     &  75.9    & 0.9337 \\
    7    &   24.53    &  0.8038     &   0.9124    &  77.9     & 0.9246 \\
    \bottomrule
    \end{tabular}%
  \label{tab:addlabel}%
  \vspace{-0.3cm}
\end{table}%

\vspace{-0.2cm}
\subsubsection{The network depth}
Finally, we explored the influence of the Transformer blocks number. The results are shown in Table \ref{different depth}. When the number of blocks is 5, the performance is optimal. To a certain extent, the performance improvement is not apparent when increasing the block number. This phenomenon means that increasing the number of blocks does not improve performance indefinitely.
\section{Conclusion}
In this paper, we design a network for correcting fisheye images with the help of Transformer structure, processing the fisheye images as a sequence for the first time. Considering the Transformer slicing method is not appropriate, an annulus slicing that conforms to the distortion distribution of the fisheye image is proposed. According to the observation of the parameters, we found the distortion efficacy domain and designed a novel layer attention mechanism to enhance the local texture perception. For the problem of inconsistency confidence of the different patch parameters, we introduce a set of truncated normal distribution probability densities as the weight of patch loss to make the training more reasonable. Our network has a structure that combines global and local perceptions. The test results on two completely different datasets(Place2 and CelebA) prove the effectiveness of our method. Cross-testing between models further verifies that our approach has a better performance.

%
%
\bibliographystyle{unsrt}
\bibliographystyle{splncs04}
\bibliography{ref}

\end{document}